\documentclass[sn-mathphys,Numbered]{sn-jnl}% Math and Physical Sciences Reference Style
\DeclareUnicodeCharacter{2212}{-}
\usepackage{graphicx}%
\usepackage{multirow}%
\usepackage{amsmath,amssymb,amsfonts}%
\usepackage{amsthm}%
\usepackage{mathrsfs}%
\usepackage[title]{appendix}%
\usepackage{xcolor}%
\usepackage{textcomp}%
\usepackage{manyfoot}%
\usepackage{booktabs}%
\usepackage{algorithm}%
\usepackage{algorithmicx}%
\usepackage{algpseudocode}%
\usepackage{listings}%
\usepackage[symbol]{footmisc}

\theoremstyle{thmstyleone}%
\theoremstyle{thmstyletwo}%

\theoremstyle{thmstylethree}%

\begin{document}

\title[Article Title]
{Shifting to Machine Supervision: Annotation-Efficient Semi and Self-Supervised Learning for Automatic Medical Image Segmentation and Classification}

%{Shifting to Machine Supervision: Enhancing Medical Image Segmentation and Classification through Semi and Self-Supervised Learning}

% \title[Article Title]
% {Data efficient pipelines for medical image analysis}
%{Improving Medical Image Segmentation and Classification with limited human supervision}

%Improving Medical Image Segmentation with increased machine supervision and reduced human supervision}

%%=============================================================%%
%% Prefix	-> \pfx{Dr}
%% GivenName	-> \fnm{Joergen W.}
%% Particle	-> \spfx{van der} -> surname prefix
%% FamilyName	-> \sur{Ploeg}
%% Suffix	-> \sfx{IV}
%% NatureName	-> \tanm{Poet Laureate} -> Title after name
%% Degrees	-> \dgr{MSc, PhD}
%% \author*[1,2]{\pfx{Dr} \fnm{Joergen W.} \spfx{van der} \sur{Ploeg} \sfx{IV} \tanm{Poet Laureate} 
%%                 \dgr{MSc, PhD}}\email{iauthor@gmail.com}
%%=============================================================%%

\author[1]{\fnm{Pranav} \sur{Singh}}\email{ps4364@nyu.edu}

\author[2]{\fnm{Raviteja} \sur{Chukkapalli}}\email{rc5124@nyu.edu}
%\equalcont{These authors contributed equally to this work.}

\author[2]{\fnm{Shravan} \sur{Chaudhari}}\email{shravan.c@nyu.edu}
%\equalcont{These authors contributed equally to this work.}

\author[1]{\fnm{Luoyao} \sur{Chen}}\email{lc4866@nyu.edu}
%\equalcont{These authors contributed equally to this work.}

\author[1]{\fnm{Mei} \sur{Chen}}\email{mc8895@nyu.edu}
%\equalcont{These authors contributed equally to this work.}

\author[1]{\fnm{Jinqian} \sur{Pan}}\email{jp6218@nyu.edu}
%\equalcont{These authors contributed equally to this work.}
\author[3]{\fnm{Craig} \sur{Smuda}}\email{Craig.Smuda@nyulangone.org}
\author*[1,4]{\fnm{Jacopo} \sur{Cirrone}}\email{cirrone@courant.nyu.edu}
%\equalcont{These authors contributed equally to this work.}

\affil[1]{\orgdiv{Center for Data Science}, \orgname{New York University}, \orgaddress{\street{60 5th Ave}, \city{New York}, \postcode{10011}, \state{NY}, \country{USA}}}

\affil[2]{\orgdiv{Department of Computer Science, Courant Institute of Mathematical Sciences}, \orgname{New York University}, \orgaddress{\street{251 Mercer St}, \city{New York}, \postcode{10012}, \state{NY}, \country{USA}}}

\affil[3]{\orgdiv{Division of Rheumatology}, \orgname{NYU Grossman School of Medicine}, \orgaddress{\street{550 1st Ave}, \city{New York}, \postcode{10016}, \state{NY}, \country{USA}}}

\affil[4]{\orgdiv{Colton Center for Autoimmunity}, \orgname{NYU Grossman School of Medicine, Science Building}, \orgaddress{\street{435 E 30th St}, \city{New York}, \postcode{10016}, \state{NY}, \country{USA}}}

\abstract{
Advancements in clinical treatment are increasingly constrained by the limitations of supervised learning techniques, which depend heavily on large volumes of annotated data. The annotation process is not only costly but also demands substantial time from clinical specialists. Addressing this issue, we introduce the S4MI (Self-Supervision and Semi-Supervision for Medical Imaging) pipeline, a novel approach that leverages advancements in self-supervised and semi-supervised learning. These techniques engage in auxiliary tasks that do not require labeling, thus simplifying the scaling of machine supervision compared to fully-supervised methods. Our study benchmarks these techniques on three distinct medical imaging datasets to evaluate their effectiveness in classification and segmentation tasks. Notably, we observed that self-supervised learning significantly surpassed the performance of supervised methods in the classification of all evaluated datasets. Remarkably, the semi-supervised approach demonstrated superior outcomes in segmentation, outperforming fully-supervised methods while using 50\% fewer labels across all datasets. In line with our commitment to contributing to the scientific community, we have made the S4MI code openly accessible, allowing for broader application and further development of these methods. The code can be accessed at {\href{https://github.com/pranavsinghps1/S4MI}{link}}.

}


%%\pacs[JEL Classification]{D8, H51}

%%\pacs[MSC Classification]{35A01, 65L10, 65L12, 65L20, 65L70}

\maketitle

\section{Introduction}\label{sec1}
Medical imaging analysis plays a pivotal role in clinical decision-making, aiding in diagnosis, treatment planning, and monitoring. The advent of deep learning has significantly enhanced the capability to analyze medical images both effectively and efficiently, promising to automate aspects of the diagnostic process and thereby augment clinical decision-making. However, the efficacy of these deep learning techniques is heavily contingent upon the availability of well-annotated medical image data. Unlike in natural image processing, obtaining annotations in the medical domain is fraught with challenges due to privacy concerns, high costs, and the extensive time required for expert clinicians to produce accurate annotations \cite{matsoukas2022makes}.
Moreover, the unique requirements for expert knowledge in medical imaging mean that traditional crowd-sourcing for annotations is not viable. This results in medical imaging datasets being significantly smaller than their natural imaging counterparts, leading to suboptimal performance when neural networks are trained from scratch on such limited data volumes. Consequently, transfer learning has become a key strategy, leveraging knowledge acquired from source domain tasks to enhance performance on target domain tasks in medical imaging.
Transfer learning typically involves initializing the network with weights from a pre-trained model and subsequently fine-tuning it with target domain data. While transfer learning has shown promise, especially when source and target datasets, as well as their respective output classes, are similar \cite{yang2020transfer}, the distinct nature of medical imaging often limits its applicability.
\\
The core motivation of our study is to address these challenges by comparing machine supervision with traditional fully-supervised approaches in the realm of medical imaging. The acquisition of labels in medical imaging is notably costly and time-intensive, a hurdle that machine supervision approaches, with their scalable and label-free nature, aim to overcome. By enabling the development of priors that can outperform supervised methods, machine supervision tackles a critical bottleneck in advancing clinical treatments—the heavy reliance on supervised learning techniques that necessitate extensive annotated data. This study aims to elucidate the efficiencies that machine supervision can introduce, particularly in light of the challenges posed by the need for domain-specific annotation, the limited size of medical datasets due to stringent privacy regulations and high annotation costs, and the indispensable requirement for expert knowledge in annotating medical images.
\\

In recent times, the field of computer vision has witnessed remarkable strides due to achievements in learning algorithms such as Deep Metric Learning \cite{chen2020simple}, self-distillation \cite{caron2021emerging}, masked image modeling \cite{xie2022simmim}, etc. These innovative approaches have enabled the learning of meaningful visual representations from unannotated image data, with fine-tuning models from these learners showing competitive performance gains \cite{caron2021emerging}. There is a growing interest in exploring and applying these learning algorithms to medical imaging modalities, where annotations are notably scarce. These algorithms, typically deployed in self-supervised, semi-supervised, or unsupervised settings, involve a two-stage process: 1) pre-training to acquire representations from unannotated data, and 2) fine-tuning using annotated data to refine the model for the specific target task.

In the realm of self-supervised learning, our work specifically focuses on DINO (Distillation with NO labels) \cite{caron2021emerging} and CASS (Cross-Architectural Self-supervision) \cite{singh2022cass} techniques, which are at the forefront of current research. These methods leverage joint embedding \cite{balestriero2023cookbook} based architecture, a cutting-edge approach that facilitates learning meaningful visual representations by aligning two different views of an image to the same embedding space. This strategy is particularly adept at extracting robust features from unannotated data, making it highly relevant to medical imaging analysis.
Additionally, for semi-supervised learning, we delve into the cross-architectural, cross-teaching method \cite{luo2021ctbct}, which represents a significant advancement towards utilizing machine supervision over traditional human annotation. The adoption of such state-of-the-art self/semi-supervised techniques marks a pivotal shift in medical imaging, emphasizing machine supervision's scalable and label-free advantages. This is crucial in the context of medical imaging, where the procurement of labels is not only cost-prohibitive but also requires extensive time and expertise. 
\\
Building upon these considerations, this work introduces the S4MI (Self-Supervision and Semi-Supervision for Medical Imaging) pipeline. This novel semi-/self-supervised learning approach is designed to directly address the aforementioned challenges by minimizing the dependency on extensive labeling efforts. By integrating cutting-edge self-supervised learning algorithms with the latest semi-supervised learning techniques, S4MI aims to significantly improve performance beyond the capabilities of traditional supervised approaches. This initiative represents a significant step towards the development of scalable healthcare solutions, leveraging the untapped potential of machine supervision to achieve and potentially exceed the efficacy of human-supervised methods in medical imaging analysis.
\\
In this work, we evaluate the performance of state-of-the-art machine supervision approaches against traditional transfer learning across two pivotal tasks in medical image analysis: segmentation and classification. Our comparison spans three challenging datasets from distinct medical modalities, including histopathology slide images and skin lesion images. Additionally, we explore the effectiveness of machine supervision by varying the amount of annotated data used for fine-tuning. The machine supervision approaches examined in this study include \cite{caron2021emerging, singh2022cass} for classification and \cite{luo2021ctbct, picie2021} for segmentation.

\section{Data}
\label{dataset}
To compare transfer learning from ImageNet with machine supervision and show its general applicability, we selected three datasets representative of those clinicians encounter in real life, featuring varying levels of class imbalance and sample sizes.
%To compare transfer learning from ImageNet with machine supervision, we select three datasets representative of the datasets clinicians have to deal with in real life, with varying levels class imbalance and sample sizes.

\begin{itemize}
  \item \textbf{Dermatomyositis}: 
  This dataset comprises 198 RGB samples from 7 patients, each image measuring 352 by 469 pixels. With this dataset, we conduct multilabel classification, aiming to classify cells based on their protein staining into TFH-1, TFH-217, TFH-Like, B cells, and cells that do not conform to the aforementioned categories, labeled as 'others.' We utilize the F1 score as our evaluation metric on the test set.
  %We have 198 RGB samples from 7 patients, each of 352 by 469 in size. With this dataset, we perform multilabel classification. Our task here is to classify cells based on their protein staining into TFH-1, TFH-217, TFH-Like, B cells, and cells that don't conform to previous cell types as ``others". We use the F1 score as our evaluation metric on the test set. 
  \item \textbf{Dermofit}: This dataset consists of 1,300 image samples captured with an SLR camera, spanning the following ten classes: Actinic Keratosis (AK), Basal Cell Carcinoma (BCC), Melanocytic Nevus / Mole (ML), Squamous Cell Carcinoma (SCC), Seborrhoeic Keratosis (SK), Intraepithelial Carcinoma (IEC), Pyogenic Granuloma (PYO), Haemangioma (VASC), Dermatofibroma (DF), and Melanoma (MEL). Each image in this dataset is uniquely sized, with dimensions ranging from 205×205 to 1020×1020 pixels.
  %This dataset contains 1300 image samples taken using an SLR camera with the following ten classes: Actinic Keratosis (AK), Basal Cell Carcinoma (BCC), Melanocytic Nevus / Mole (ML), Squamous Cell Carcinoma (SCC), Seborrhoeic Keratosis (SK), Intraepithelial carcinoma (IEC), Pyogenic Granuloma (PYO), Haemangioma (VASC), Dermatofibroma (DF) and Melanoma (MEL). No two images in this dataset are the same size, varying from 205×205 to 1020×1020.
  \item \textbf{ISIC-2017}: 
  Within this dataset, 2,000 JPEG images are distributed among three classes: Melanoma, Seborrheic Keratosis, and Benign Nevi. The evaluation of the test set is based on the Recall score.
  %For this dataset, we have 2000 JPEG images classified into three classes - Melanoma, Seborrheic keratosis, and Benign nevi. We use the Recall score as our evaluation metric on the test set. 
\end{itemize}

\section{Methods}\label{sec11}

\subsection{Classification}  
\label{classification_method}

For classification, we compare existing self-supervised techniques with transfer learning, the de facto norm in medical imaging. As mentioned, due to a lack of data in medical imaging, classifiers are often trained using initialization from other large visual datasets like ImageNet \cite{deng2009imagenet}. The architectures are then fine-tuned on the target medical imaging dataset using these initializations. Alternatively, we can train using self-supervised approaches to provide better classification priors for the target medical imaging dataset. In self-supervised learning, an auxiliary task is performed without labels to learn fine-grained information about the image. This auxiliary task can be performed in various ways, for example, by corrupting the image, followed by its reconstruction, creating copies of the same image (positive pairs), and minimizing the distance between them using two differently parameterized architectures or redundancy reduction. In our study, we focus on DINO \cite{caron2021emerging} and CASS \cite{singh2022cass}. DINO learns by creating augmentation-asymmetric copies of the input image, whereas CASS creates architecturally asymmetric copies followed by similarity maximization between the copies. We pre-trained using these two self-supervised techniques for 100 epochs, followed by 50 epochs of fine-tuning with labels. We perform fine-tuning with 10\% and 100\% label fractions. For this, we use all of the available labels per image while using 10\% or 100\% of the total labels available.

\subsubsection{Classification pipeline additional details} 

For the ImageNet supervised classification approach, we used images and their corresponding labels as our inputs. For training with x\% labels, we only used x\% of the entire training images and labels while keeping the number of labels per image the same. For the self-supervised approaches DINO and CASS, for pertaining, we start with only images. During the fine-tuning process, we initialize the networks with their pre-trained weights and use corresponding image-label pairs. Similar to the supervised approach, we also fine-tune these architectures for two label fractions, 10\% and 100\%. Further details can be inferred from our open-sourced \href{https://github.com/pranavsinghps1/S4MI.}{code base}.

\subsection{Segmentation}            
% Describe the experimental methodology that you used. What are the criteria that you are using to evaluate your method? What specific hypotheses does your experiment test? If relevant for your project, did you do training/validate/test splits? Comparisons to competing methods that address the same problem are particularly useful, if relevant.   

% training/validate/test splits
% Data Augmentation:
% ToPILImage
% Flip, rotation
% Normalization: Dermofit: Rnorm (why not parameter from ImgNet)
% ToTensor
% Resize

% optimizer = optim.Adam(model.parameters(), lr=3.6e-04, weight_decay=1e-05)
% optimizer_ae = optim.Adam(autoencoder.parameters(), lr=1e-3)
% scheduler = optim.lr_scheduler.CosineAnnealingLR(optimizer, 50, eta_min=3.4e-04, last_epoch=-1)
This section will first introduce standard implementation designs mutually applied to all models, followed by additional implementation details for the semi-supervised model. Lastly, we will describe the procedure of testing different cross-entropy weight initialization.

\subsubsection{Approach specific additional details}
\paragraph{Semi-Supervised Approach :} The semi-supervised model uses a batch size of 16, and each batch consists of eight labeled images and eight unlabeled images. In the last column of Figure \ref{table: fullyvsSemi}, we compare the performances between semi- and fully-supervised models in 100\% labeled-ratio scenario; for this, we re-adjusted the batch to be 15 labeled images and one unlabeled image to approximate the fully-supervised setting while still retaining the ``cross teaching” component so that the loss remains consistent and hence comparable with other semi-supervised label ratios. Additionally, when comparing the performance between fully and semi-supervised models, we adopt the same practice from \citet{luo2021ctbct} and use Swin Transformer to compare DEDL \cite{singh2023data} with Resnet34 backbone as they have a similar number of parameters

\paragraph{Unsupervised Approach :} 
We implemented PiCIE for Dermofit, Dermatomyositis, and ISIC-2017 datasets similar to the supervised and self/semi-supervised methods. We made slight changes that suit the datasets and ensure smooth learning, such as using SGD optimizer instead of Adam optimizer, keeping the learning rate the same as the original implementation (1e-4), and adding the StepLR (Step Learning Rate) scheduler provided by PyTorch. This is because we observed that the model sometimes failed to learn and was stuck at predicting all the image pixels as either foreground (thing) or background (stuff) with the Adam optimizer. We used batch sizes of 64 and 128, depending on the availability of CUDA RAM and dataset sizes. We trained the PiCIE unsupervised pipeline for 50 epochs with the ResNet34 backbone as a feature extractor and retained the unsupervised clustering technique described in \cite{picie2021} to achieve optimal segmentation performance. The hardware used was Nvidia RTX8000, similar to the previous methods. For PiCIE, labels were only used for validation and testing, not during training.

\paragraph{Supervised Approach :} 
We adopted \citet{singh2022dataefficient}'s approach for the supervised part of our comparison. For this, we utilize image and mask pairs during training, validation, and testing with a ResNet-34-based U-Net\cite{ronneberger2015u}.

\subsection{Common Implementations:}
We implemented all models in Pytorch \cite{NEURIPS2019_9015} using a single NVIDIA RTX-8000 GPU with 64 GB RAM and 3 CPU cores. All models are trained with an Adam optimizer with an initial learning rate (lr) of 3.6e-4 and a weight decay of 1e-5.
We also set a cosine annealing scheduler with a maximum of 50 iterations and a minimum learning rate of 3.4e-4 to adjust the learning rate based on each epoch.
%train test split
For splitting our dataset into training, validation, and testing sets, we use a random train-validation-test split (70\%-10\%-20\%), except in the ISIC2017 dataset, where we adopt the train/val/test split according to \cite{codella2018skin} for match-up comparison (57\%-8.5\%-34\%).
% data augmentation
The batch size is 16, and we use data augmentation to enrich the training set using random rotation, random flip, and a further resizing to 224 $\times$ 224 to fit in Swin Transformer's patch size requirement\cite{liu2021swin}. Note that for 3-channel datasets (Dermofit, ISIC), we add a pre-processing step that normalizes the red channel of the RGB color model as proposed by \citet{rnorm}. 
% metrics
%Following \citet{yu2019uncertainty, luo2021ctbct}, we use four metrics to quantitatively evaluate our model's performance, including Jaccard, Dice, 95\% Hausdorff Distance and average surface distance. Our final metrics presented in \ref{result} are all in Jaccard. 
We repeat all experiments with different seed values five times and report the mean value in the 95\% confidence interval in all tables. Similar to classification, we fine-tune with multiple label fractions, with semi-supervised and DEDL. When we mention that we fine-tune with x\% labels, we use all labels per image but only x\% of the total available image-label pairs. 

\subsubsection{Data-Preprocessing}\label{preprocessing}
Following \citet{singh2023data}, we chose the processed image sample size to be 480 × 480 to avoid empty tile issues. To ensure uniformity, the image input size for the model remains consistent across all three datasets, as illustrated below.
For the Dermatomyositis dataset, since it contains images of uniform sizes 1408$\times$1876 each, we tiled them to a size of 480$\times$480 and then used blank padding at the edges to make them fit in 480$\times$480 sizes. This results in 12 tiled sub-images per sample, which are then resized to 224$\times$224.
In contrast, the Dermofit and the ISIC2017 datasets contain images of different sizes. Since the two datasets are about skin lesions, they have significantly denser and larger mask labels than the Dermatomyositis dataset. Thus, a different image preprocessing step is applied to the latter two datasets: bilinear interpolation to 480 × 480 followed by a resize to 224 $\times$224.
\section{Results}\label{sec2}

We start by studying the classification pipeline in Section \ref{classification}, followed by understanding the interpretability aspect of classification in Section \ref{saliency}. Since interpretability is ingrained in the segmentation task, we only study segmentation quantitatively in Section \ref{subsec1}.

\subsection{Classification}\label{classification}
%\subsection{Studying and benchmarking supervised and self-supervised techniques for Classification}\label{classification}
% In this section, we benchmark and study the performance of transfer learning as well as two self-supervised techniques -- Distillation with No Labels (DINO) \cite{caron2021emerging} and Cross Architectural Self-Supervision (CASS) \cite{singh2022cass}. We introduce them in Section \ref{ssl-sec}, followed by results on three classification datasets in Table \ref{perf}.

For classification, we advocate using self-supervision to learn useful discriminative representations from unlabelled data. Self-supervised techniques rely on an auxiliary pretext task for pre-training to learn useful representations. These representations are further improved and aligned with the downstream task through labeled fine-tuning. We benchmark and study the performance of transfer learning as well as two self-supervised techniques: 1) Distillation with No Labels (DINO) \cite{caron2021emerging}, a self-supervised approach designed for natural images, and 2) Cross cross-architectural self-supervision (CASS) \cite{singh2022cass} a self-supervision approach for medical imaging. We further expand on our methodology in Section \ref{classification_method}.\\

\begin{figure}[!ht]
    \centering
    \includegraphics[width=1\textwidth]{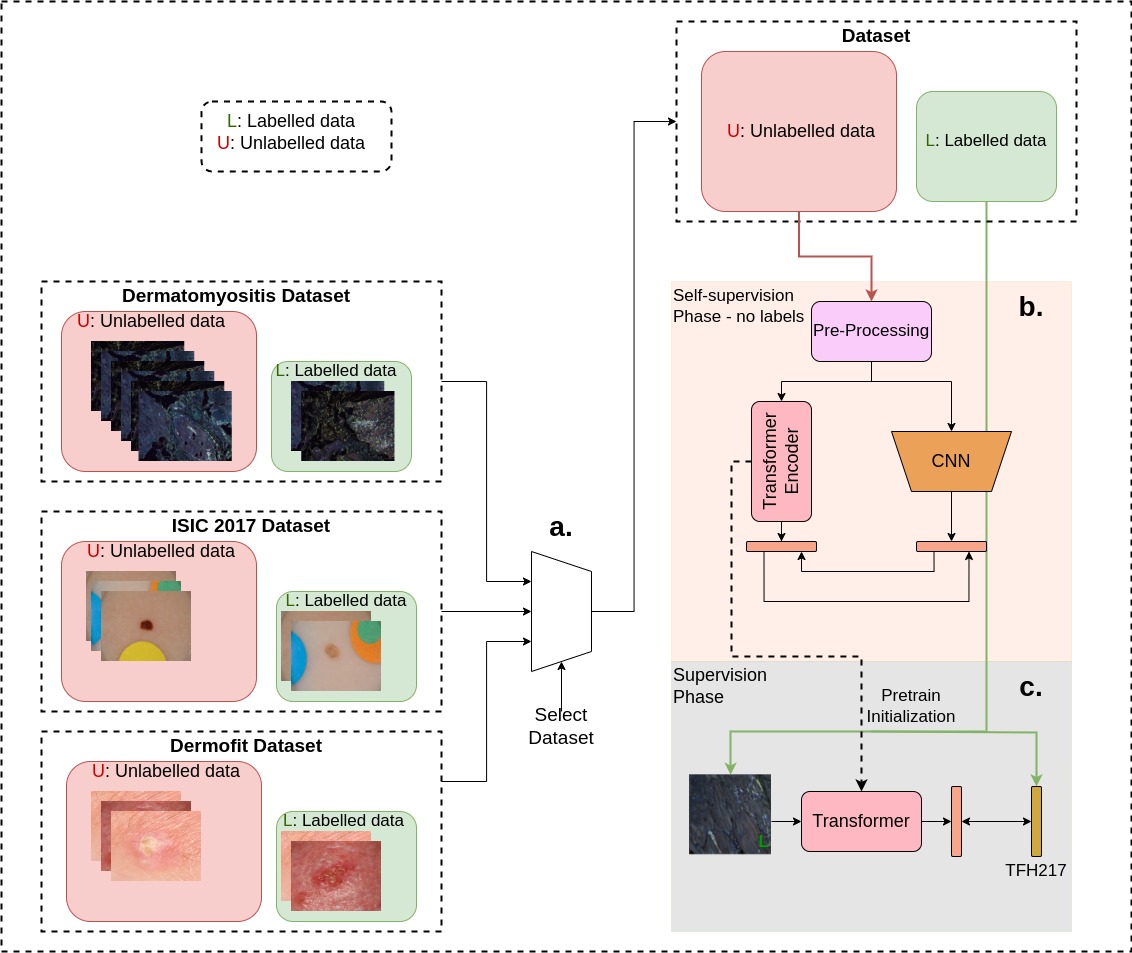}
    \caption{\textbf{Applying CASS:} In this figure we detail the steps involved in applying CASS, the best-performing machine supervision approach for classification \cite{singh2022cass}. We conducted experiments on three datasets listed on the left side: the Dermatomyositis, ISIC-2017, and Dermofit datasets. For training on a dataset, we initialize both the networks with their ImageNet weights and select one dataset at a time. To train CASS, we start by label-free pretraining as illustrated in Part (b) of Fig. \ref{fig:cls_pipeline}. During pre-training, a CNN and a Transformer are trained simultaneously. In the case of the Dermatomyositis dataset, the finetuning is multi-label, while in the case of the ISIC-2017 and the Dermofit dataset, it is multi-class. This pre-training in (b) is followed by labeled fine-tuning as shown in Part (c), where image are fine-tuned one at a time.}
    
    %We conducted experiments on three datasets - the Dermatomyositis, ISIC-2017, and the Dermofit datasets. For training on a dataset, we initialize both the networks with their ImageNet weights and select one dataset at a time. To train CASS, we start by label-free pretraining as illustrated in Part (b) of Fig. . This is followed by labeled fine-tuning as shown in Part (c). During, pre-training a CNN and a Transformer are trained simultaneously.  but during finetuning, the pre-trained architectures are finetuned one at a time. In the case of the Dermatomyositis dataset, the finetuning is multi-label while in the case of the ISIC-2017 and the Dermofit dataset, it is multi-class.}
    %The experiments are conducted over three diverse datasets as shown in the figure. For each dataset selected(a.) The color-coded top half(b.) shows the self-supervised approach used for learning useful representations from unannotated data. An intuitive way to understand the self-supervised phase(b.) is that the two different network architectures(Transformer, CNN) have different biases/priors. e.g. translational invariance is encoded into CNN but not into Transformer, this makes learning useful representations possible. The color-coded bottom half(c.) shows using annotated data to align the learnt representations to the target task through supervised fine-tuning.
    \label{fig:cls_pipeline}
\end{figure}

We present the results of supervised and self-supervised trained classifiers in Table \ref{perf}. We use the F1 score as the comparison metric, similar to its implementation in previous works \cite{singh2022cass}. The F1 score is defined as $F1 = \frac{2*Precision*Recall}{Precision+Recall} = \frac{2*TP}{2*TP+FP+FN}$, where $TP$ represents True Positive, $FN$ is False Negative and $FP$ is False Positive. We observe that CASS (Figure \ref{fig:cls_pipeline}) outperforms DINO and transfer learning consistently across all the datasets and backbones, with the, exception for the ISIC 2017 challenge dataset, where DINO outperforms CASS using the ResNet50 backbone. Interestingly, we also observe that ViT trained with CASS using just 10\% labels, which saves significant annotation time, performs significantly better than the transfer learning-based supervised approach with 100\% labels for the Dermatomyositis dataset. For the same case, DINO performs on par with the supervised approach while using 90\% fewer labels. This shows the impact that machine supervision can have on improving access to classified medical images. DINO and CASS take an image to create asymmetry through augmentation (DINO) and architecture (CASS). These are then passed through different parameterized feature extractors (DINO) and architectural feature extractors (CASS) to create embeddings. %from the augmented images using two differently parameterized encoders.
Since the two embeddings are generated from the same image, they are expected to be similar; this is then used as the supervisory signal to maximize the similarity between the two embedding and update encoders. In CASS, architectural invariance is used to create a positive pair instead of creating augmentation invariance, as in the case of DINO. This is because CNN and Transformers learn different representations from the same image \cite{singh2022cass}. 

% Please add the following required packages to your document preamble:
% \usepackage{multirow}
\begin{table}[]
\begin{tabular}{llllllll}
\midrule
\multirow{2}{*}{Backbone}  & \multirow{2}{*}{Technique}                                  & \multicolumn{2}{l}{\begin{tabular}[c]{@{}l@{}}Autoimmune Dataset\\ F1 Score\end{tabular}}                                            & \multicolumn{2}{l}{\begin{tabular}[c]{@{}l@{}}Dermofit Dataset\\ F1 Score\end{tabular}}                                               & \multicolumn{2}{l}{\begin{tabular}[c]{@{}l@{}}ISIC 2017 Dataset\\ Recall Score\end{tabular}}                                        \\
                           &                                                             & 10\%                                                             & 100\%                                                             & 10\%                                                              & 100\%                                                             & 10\%                                                            & 100\%                                                             \\
                           \midrule
\multirow{3}{*}{ResNet-50} & DINO                                                        & \textbf{\begin{tabular}[c]{@{}l@{}}0.8237\\ ±0.001\end{tabular}} & \begin{tabular}[c]{@{}l@{}}0.84252\\ ±0.008\end{tabular}          & \begin{tabular}[c]{@{}l@{}}0.3749\\ ±0.0011\end{tabular}          & \begin{tabular}[c]{@{}l@{}}0.6775\\ ±0.0005\end{tabular}          & \begin{tabular}[c]{@{}l@{}}0.322\\ ±0.009\end{tabular}          & \textbf{\begin{tabular}[c]{@{}l@{}}0.6407\\ ±0.0363\end{tabular}} \\
                           & CASS                                                        & \begin{tabular}[c]{@{}l@{}}0.8158\\ ±0.0055\end{tabular}         & \textbf{\begin{tabular}[c]{@{}l@{}}0.8650\\ ±0.0001\end{tabular}} & \textbf{\begin{tabular}[c]{@{}l@{}}0.4367\\ ±0.0002\end{tabular}} & \textbf{\begin{tabular}[c]{@{}l@{}}0.7132\\ ±0.0003\end{tabular}} & \textbf{\begin{tabular}[c]{@{}l@{}}0.343\\ ±0.002\end{tabular}} & \begin{tabular}[c]{@{}l@{}}0.599\\ ±0.0304\end{tabular}           \\
                           & \begin{tabular}[c]{@{}l@{}}ImageNet\\ Transfer\end{tabular} & \begin{tabular}[c]{@{}l@{}}0.819\\ ±0.0216\end{tabular}          & \begin{tabular}[c]{@{}l@{}}0.83895\\ ±0.007\end{tabular}          & \begin{tabular}[c]{@{}l@{}}0.33\\ ±0.0001\end{tabular}            & \begin{tabular}[c]{@{}l@{}}0.6341\\ ±0.0077\end{tabular}          & \begin{tabular}[c]{@{}l@{}}0.288\\ ±0.091\end{tabular}          & \begin{tabular}[c]{@{}l@{}}0.5774\\ ±0.0004\end{tabular}          \\
                           \midrule
\multirow{3}{*}{ViT B16}   & DINO                                                        & \begin{tabular}[c]{@{}l@{}}0.8445\\ ±0.0008\end{tabular}         & \begin{tabular}[c]{@{}l@{}}0.8639\\ ±0.002\end{tabular}           & \begin{tabular}[c]{@{}l@{}}0.332\\ ± 0.0002\end{tabular}          & \textbf{\begin{tabular}[c]{@{}l@{}}0.4810\\ ±0.0012\end{tabular}} & \begin{tabular}[c]{@{}l@{}}0.291\\ ±0.005\end{tabular}          & \begin{tabular}[c]{@{}l@{}}0.5571\\ ± 0.076\end{tabular}          \\
                           & CASS                                                        & \textbf{\begin{tabular}[c]{@{}l@{}}0.8717\\ ±0.005\end{tabular}} & \textbf{\begin{tabular}[c]{@{}l@{}}0.8894\\ ±0.005\end{tabular}}  & \textbf{\begin{tabular}[c]{@{}l@{}}0.3896\\ ±0.0013\end{tabular}} & \begin{tabular}[c]{@{}l@{}}0.4667\\ ±0.0002\end{tabular}          & \textbf{\begin{tabular}[c]{@{}l@{}}0.321\\ ±0.007\end{tabular}} & \textbf{\begin{tabular}[c]{@{}l@{}}0.588\\ ±0.0094\end{tabular}}  \\
                           & \begin{tabular}[c]{@{}l@{}}ImageNet\\ Transfer\end{tabular} & \begin{tabular}[c]{@{}l@{}}0.8356\\ ±0.007\end{tabular}          & \begin{tabular}[c]{@{}l@{}}0.8420\\ ±0.009\end{tabular}           & \begin{tabular}[c]{@{}l@{}}0.299\\ ±0.002\end{tabular}            & \begin{tabular}[c]{@{}l@{}}0.456\\ ±0.0077\end{tabular}           & \begin{tabular}[c]{@{}l@{}}0.277\\ ±0.007\end{tabular}          & \begin{tabular}[c]{@{}l@{}}0.5322\\ ±0.022\end{tabular}  \\
\midrule

\end{tabular}
\caption{Results for DINO, CASS, and Supervised (ImageNet-initialized transfer learning) methods on the Dermatomyositis, the Dermofit, and ISIC 2017 datasets. In this table, we compare the F1 score on the test sets across different backbones and label percentages. Bold values indicate the highest F1 score achieved by any method (DINO, CASS, or ImageNet transfer learning) for each combination of backbone, dataset, and label percentage. In all cases, self-supervised pre-training (CASS or DINO) improves performance over the ImageNet Transfer learning benchmark. Additionally, among CASS and DINO, we observe that CASS, with its focus on architectural invariance, outperforms DINO for almost all label fractions across three datasets for ResNet as well as for ViT.}
\label{perf}
\end{table}

\subsubsection{Saliency Maps}
\label{saliency}
\begin{figure}[!h]
    \centering
    \includegraphics[width=0.9\textwidth]{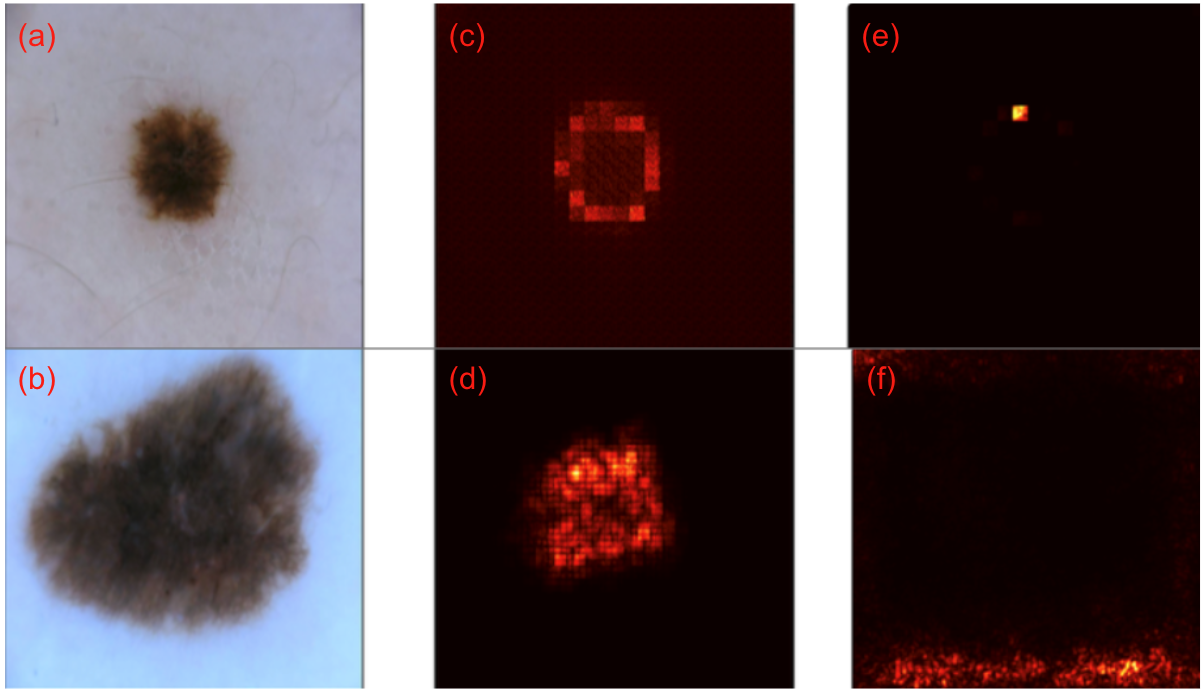} \\
\caption{Visualization of saliency maps on a random sample from the ISIC-2017 dataset, left (a, b): data (input image), middle (c, d): saliency map from CASS, and right (e,f): saliency map from DINO 
with ViTB/16 at the top and ResNet-50 at the bottom. DINO’s saliency map exhibits notable stochasticity, displaying a lack of strong correlation with the specific pathology under consideration. Conversely, in the case of CASS, the saliency maps demonstrate a significantly more aligned with the pathology of interest both for CNN as well as the Transformer.}%hile we observe that the saliency map of DINO is highly random and not hinged on the pathology of interest, for CASS they are much more aligned with the pathology of interest.}

\label{fig:Dermatomyositis_r50_1}
\end{figure}

In segmentation, interpretability is ingrained in the task itself, as we can easily identify whether model predictions are aligned with the ground truth.
To understand the decision-making process of trained neural classifiers, we study pixel attribution or saliency maps \cite{selvaraju2022grad}. To accomplish this, we first train the neural network and compute the gradient of a class score with respect to the input pixels. Backpropagating the gradients that lead to a particular classification onto the input image helps us understand the most filtered/rewarding extracted features related to classification.
%This gives an understanding of what part the classifier can classify a particular class; using this, we can also see if the neural network is paying adequate attention to all the images or just certain parts. 
We present the results using DINO, CASS-trained ResNet-50, and ViT Base/16 on the ISIC-2017 dataset in Figure \ref{fig:Dermatomyositis_r50_1}. Although most saliency maps fail to align with human-interested pathology, CASS-based saliency maps align more than those from DINO. This could be attributed to the joint training of CASS, where it trains a CNN that focuses on local information while a Transformer focuses on global or image-level features. Although both techniques fail to recognize and focus on the relevant area in all cases, CASS-trained architectures are consistently better focused on the relevant pathology than DINO-trained architectures.

\subsection{Segmentation} \label{subsec1}

\begin{figure}[!ht]
    \centering
    \includegraphics[width=1\textwidth]{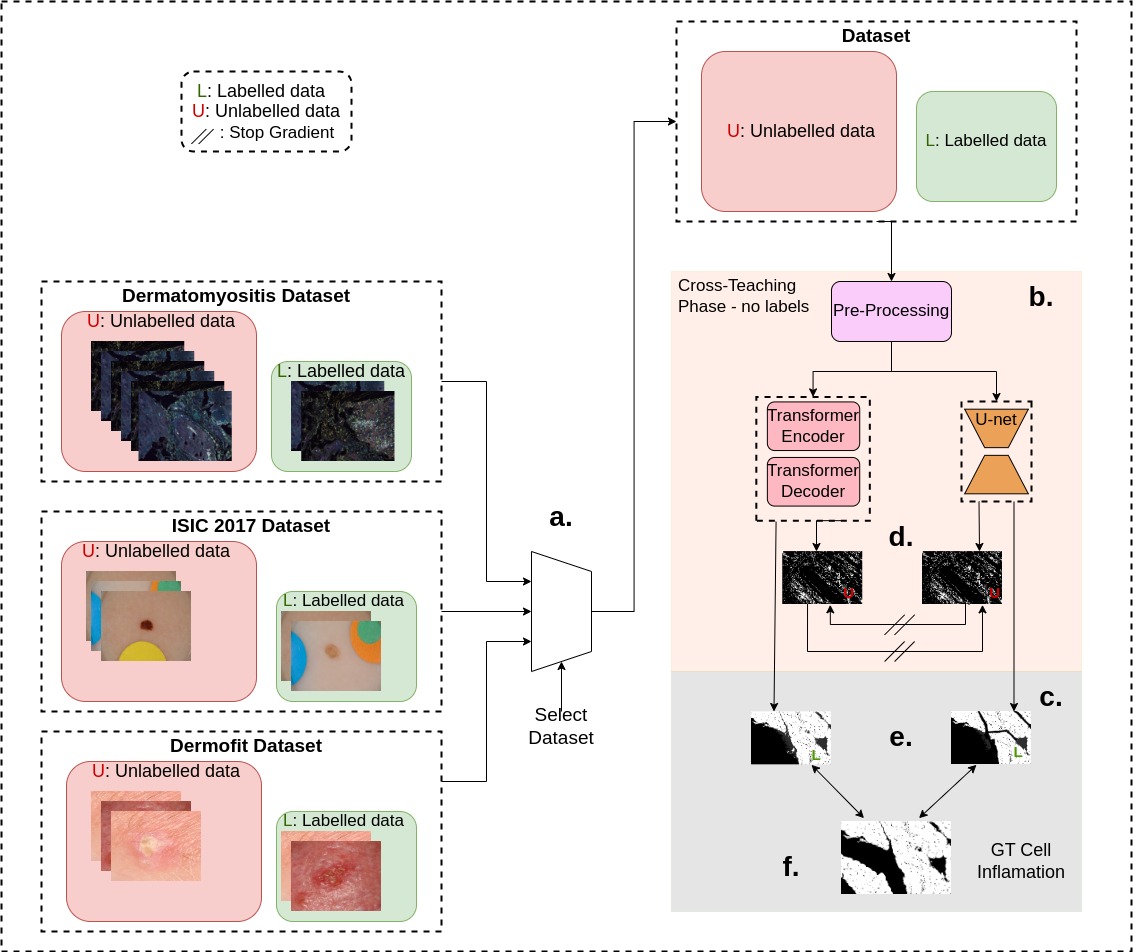}
    \caption{In this figure we present the semi-supervised pipeline as described in \ref{semi-sl-section}. Similar to the classification experiments, we evaluate the segmentation pipeline on three challenging medical image segmentation datasets - the Dermatomyositis, ISIC-2017, and the Dermofit dataset. We use one dataset at a time to train the semi-supervised architecture. Unlike the classification pipeline, semi-supervised learning involves simultaneous learning from labeled and unlabeled data. In Part (b) of Fig. \ref{fig:seg_pipeline}, we start by training the data in an unlabeled fashion and during the same iteration labeled data is also passed to the architecture as shown in part (c) of the figure. Predictions from passing inputs of the labeled images yield learned predictions as shown in Part (e). Unsupervised loss is then calculated by comparing the outputs of the CNN and the Transformer (as shown in Part(d) of this figure) using the $\mathcal{L}_{\text{Unsupervised}}$ in Section \ref{semi-sl-section}. This unsupervised loss is then added to the supervised loss denoted by $\mathcal{L}_{\text{Supervised}}$in Section \ref{semi-sl-section}. The supervised loss is calculated against the ground truth as shown in part (e) and (f) in this figure.}
    \label{fig:seg_pipeline}
\end{figure}

% \subsection{Segmentation Objective}
Image Segmentation is another important task in medical image analysis. The area of interest -- usually containing the pathology -- is segmented from the rest of the slide image. We study the performance of transfer learning, semi-supervised, and unsupervised approaches for segmentation. In computer vision, it is widely acknowledged that image segmentation presents a more intricate challenge than image classification. This distinction arises from the fact that image segmentation necessitates the meticulous classification of individual pixels, whereas image classification solely requires making predictions at the image level. Moreover, due to the aforementioned rationale, annotating images for segmentation exhibits a significantly higher complexity level. Therefore, the present study aims to investigate the effectiveness of unsupervised, semi-supervised, and fully supervised techniques across four distinct label fractions instead of the conventional two in classification. In our case, x\% label fractions indicate that we only use x\% labels from the dataset.
%Image Segmentation is another important task in medical image analysis. The area of interest -- usually containing the pathology -- is segmented from the rest of the slide image. We study the performance of transfer learning, semi-supervised, and unsupervised approaches for segmentation. In computer vision, it is widely acknowledged that image segmentation presents a more intricate challenge than image classification. This distinction arises from the fact that image segmentation necessitates the meticulous classification of individual pixels, whereas image classification solely requires making predictions at the image level. Moreover, due to the aforementioned rationale, annotating images for segmentation exhibits a significantly higher complexity level. Therefore, the present study aims to investigate the effectiveness of unsupervised, semi-supervised, and fully supervised techniques across four distinct label fractions instead of the conventional two in classification. In our case, x\% label fractions indicate that we only use x\% labels from the dataset.
We include the unsupervised approach \cite{picie2021} to study the performance for 0\% labels scenario. We compare the performance of DEDL (Data-Efficient Deep Learning framework) \cite{singh2023data}, a transfer learning-based approach, against \cite{luo2021ctbct}, a semi-supervised approach using the Swin transformer-based U-Net model (Figure \ref{fig:seg_pipeline}).
\\
We choose \cite{singh2023data} due to its significant promise in histopathology imaging, and \cite{luo2021ctbct} for its state-of-the-art approach to leveraging unannotated data in MR segmentation. The semi-supervised approach \cite{luo2021ctbct} trains two segmentation architectures - a Swin transformer-based U-Net and a CNN U-Net (as shown in Fig. \ref{fig:seg_pipeline}), while DEDL \cite{singh2023data} and unsupervised approach \cite{picie2021} only train one architecture. 
\\
To ensure a fair comparison, we make sure the number of overall parameters trained is the same in each case over three datasets and four-label fractions.
% We propose to use semi-supervised learning for learning from unannotated data alongside annotated data. 
\\
In their study, \citet{luo2021ctbct} compared the performance of Swin-U-Net trained in semi- and fully-supervised fashion for segmentation. 
Observing that Swin-U-Net exhibits superior performance in a semi-supervised setting, we extend this comparison to include a semi-supervised Swin-U-Net and a similarly parameterized ResNet-34 based U-Net, trained under both unsupervised and fully-supervised approaches. Furthermore, from \cite{singh2023data}, we use their best-reported ResNet-34-based U-Net model. 
\\
The evaluations are based on three datasets, as detailed in Section \ref{dataset}. To ensure our results are statistically significant, we conduct all experiments with five different seed values and report the mean values in the 95\% confidence interval (C.I.) over the five runs. For comparison, we use IoU (Intersection over Union) in line with previous works in the field \cite{singh2022dataefficient}. IoU or Jaccard index for two images U and V is defined as $IoU(U,V) = \frac{|U \cap V|}{|U \cup V|}$. We present these results in Figure \ref{table: fullyvsSemi}.\\
\subsubsection{Semi-Supervised Approach}
\label{semi-sl-section}
\begin{figure}[!ht]
    \centering
     %\includegraphics[width=1\textwidth]{image/Fig_3_a_alter.png}
    %\includegraphics[width=1\textwidth]{image/pipeline.png}
     %\text{(a)}
     %\includegraphics[width=0.9\textwidth]{image/Fig_3_b.png}
     \includegraphics[width=1\textwidth]{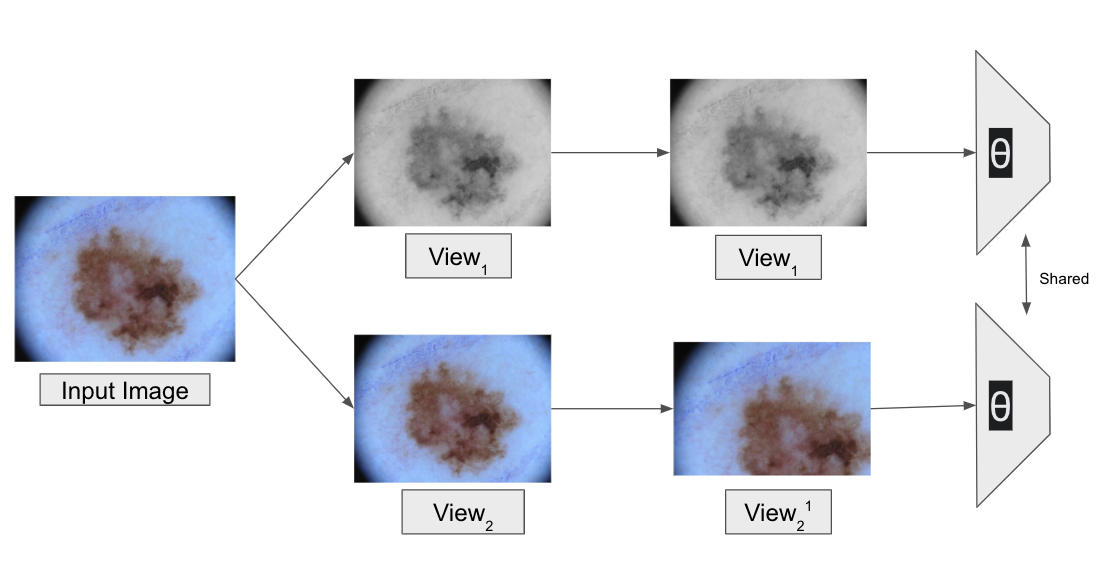}
    %\text{(b)}
    \caption{ %In part (a) of the figure, we describe the Semi-Supervised Approach: Cross Teaching Between CNN and Transformer\cite{ssl4mis2020}. Since a batch used in this approach constitutes both labeled and unlabeled portions. Here, Labeled images represent the image with corresponding ground truth labels of the batch and unlabelled parts of the batch, i.e., only images. The U-Net and the Swin U-Net receive this batch with labeled and unlabeled parts. So, they produce outputs on both parts of the dataset - labeled and unlabeled. Dice loss is calculated between the unlabeled parts, while a cross-entropy loss is calculated using the labeled parts against the ground truth. 
    In this figure, we depict the Unsupervised Segmentation Approach: PiCIE pipeline \cite{picie2021}. $View_1$ and $View_2$ represent two photometrically transformed views of the input image, whereas $View_2^{1}$ represents the geometric transformation of $View_2$. Cross-view training is then used to train the architecture shared between the two views (parameterized by $\theta$ in the figure); we have expanded further on this in Section \ref{unsup}.} % Should add more detail
    %\caption{ In this figure, we depict the Unsupervised Segmentation Approach: PiCIE pipeline \cite{picie2021}. $View_1$ and $View_2$ (at the bottom) represent two photometrically transformed views of the input image, whereas $View_1$ and $View_2$ (at the top) represent the geometric transformation of one of the views. Cross-view training is then used to train the architecture; we have expanded further on this in Section \ref{unsup}.}
    \label{fig:SSL}
\end{figure}

Semi-supervision approaches utilize the dataset's labeled and unlabeled parts for learning. Techniques generally fall under 1) Adversarial training (e.g. DAN\cite{10.1007/978-3-319-66179-7_47}) 2) Self-Training (e.g. MR image segmentation\cite{10.1007/978-3-319-66185-8_29})  3) Co-Training (e.g. DCT\cite{peng2020deep}) and 4) Knowledge Distillation (e.g. UAMT\cite{yu2019uncertainty}). We study the co-training-based semi-supervision segmentation technique \cite{luo2021ctbct}, depicted  in Figure \ref{fig:SSL}. The model consists of two segmentation architectures, a U-Net (trained from scratch) and a pre-trained Swin Transformer, adapted from the Swin Transformer proposed by Liu et al.\cite{liu2021swin}. Both architectures are updated simultaneously using a combined loss over the labeled and unlabeled images.\\

% \begin{figure}[!ht]
%     \centering
%     \includegraphics[width=1\textwidth]{image/paper_edits.drawio.png}
%     \caption{In this figure (on the right, subsections b, d, e, and f), we describe the overall process of medical image analysis using the semi-supervised segmentation pipeline. In (b), we take a batch of labeled and unlabeled images as input and pass it to the Swin-U-Net and U-Net.
%     Then, from this architecture, we get an on the unlabeled and the labeled portion - d and e, respectively.  A stop-gradient is used to stop the flow of gradients based on the loss from the unlabelled portion of the data. Passing the input image through the segmentation architecture yields a mask of the foreground (pathology of interest) and that of the background (area other than the pathology of interest). Similarly, on the left, we describe the self-supervised training process followed by the classifier fine-tuning with labeled data. For this, we first only the unlabeled in the self-supervised training (a), followed by fine-tuning with labeled data in (c).}
%     \label{fig:pipeline}
% \end{figure}

The U-Net model is trained as follows: \\
1) The \textbf{labeled data} $(\text{L})$ is evaluated using the average of dice and cross-entropy loss
$$\mathcal{L}_{\text{Supervised}} = \frac{1}{2}[\text{DICE}(\text{pred}_{\text{U-Net}}, \text{L}) + \text{CrossEntropy}(\text{pred}_{\text{U-Net}}, \text{L})]$$
2) For \textbf{unlabeled data} $(\text{U})$ cross-teaching strategy is used to cross-supervised between CNN and Transformer: to update U-Net, output from Transformer, $\text{pred}_{\text{Transformer}}$ (vice versa) is used as "sudo ground truth":
$$\text{U}_{\text{sudo}} = \text{argmax}(\text{pred}_{\text{Transformer}})$$
3) The unsupervised loss for U-Net is conducted between prediction and pseudo-ground truth:
$$\mathcal{L}_{\text{Unsupervised}} = \text{DICE}(\text{pred}_{\text{U-Net}}, \text{U}_{\text{sudo}})$$
% The argmax operation stops the back-propagation, as the backward slashes indicate. \\ 
4) The final loss is the summation of unsupervised and supervised loss.\\

\subsubsection{Unsupervised Approach}
\label{unsup}
Unsupervised approaches for learning from unlabelled images are mostly limited to curated and object-centric images; however, some recently proposed methods like \cite{picie2021} achieve semantic segmentation with the help of unsupervised clustering methods. The machine aims to discover and identify the different foreground objects and separate them from the background. In this paper, the datasets under study only contain one foreground object, which eases learning; however, unsupervised segmentation continues to be a challenging task.
\cite{picie2021} proposes a technique PiCIE (Pixel-level feature Clustering using Invariance and Equivariance) to incorporate geometric consistency to learn invariance to photometric transitions and equivariance to geometric transitions. The two main constraints here are 1) to cluster the pixels having similar appearance together with the same label for all the pixels belonging to a cluster and 2) to ensure the pixel label predictions follow the invariance and equivariance constraints mentioned earlier. With these objectives, \cite{picie2021} trains a ConvNet-based semantic segmentation model in a purely unsupervised manner (without using any pixel labels). PiCIE uses the alternating strategy between clustering the feature representations and using the cluster labels as pseudo labels to train the feature representation proposed by DeepCluster \cite{deepcluster2018}.\\

\autoref{fig:SSL} (b) shows a brief overview and an end-to-end pipeline of the PiCIE approach. For each input image, two photometric transformations, like jittering the pixel colors, etc., are randomly sampled to create $View_{1}$ and $View_2^{1}$. One of these views ($View_{1}$) is directly fed to the network, while geometric transformation is applied to the other view before feeding it to the same network ($View_2^{1}$). The network then produces the two feature representations for every pixel corresponding to the two views. Since we expect equivariance for geometric transformations, they are applied to feature representations of the original image before clustering the pixels of the feature representations. The inductive biases of invariance and equivariance to photometric and geometric transformations are introduced by aggregating the clustering losses within the same view and across the two views. Using this cross-view training strategy, the training is guided to prevent predicted pixel labels from changing with jittering of pixel colors. Still, at the same time, if the image is warped geometrically, then the warping would also be reflected in the labeling similarly \cite{picie2021}. \\

\subsubsection{Comparision}
\begin{figure}[!htbp]
\includegraphics[width=0.52\textwidth]{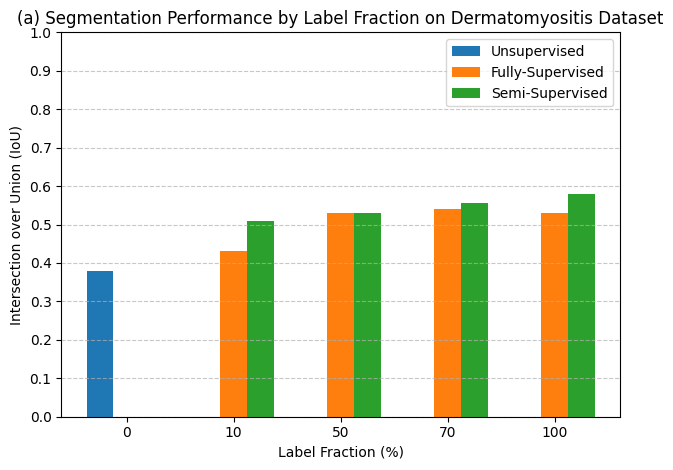}
\includegraphics[width=0.48\textwidth]{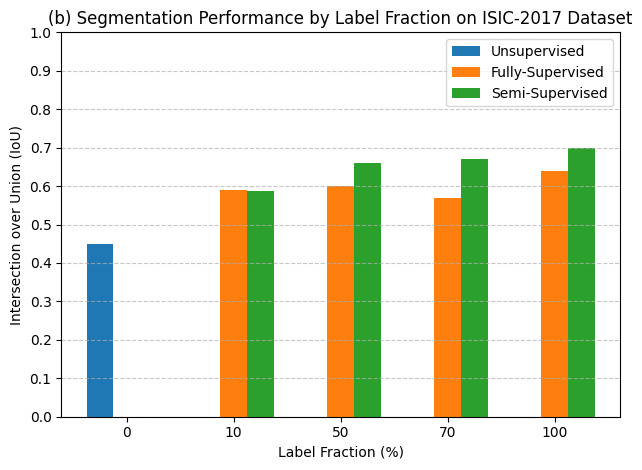}

    \centering
    \includegraphics[width=0.5\textwidth]{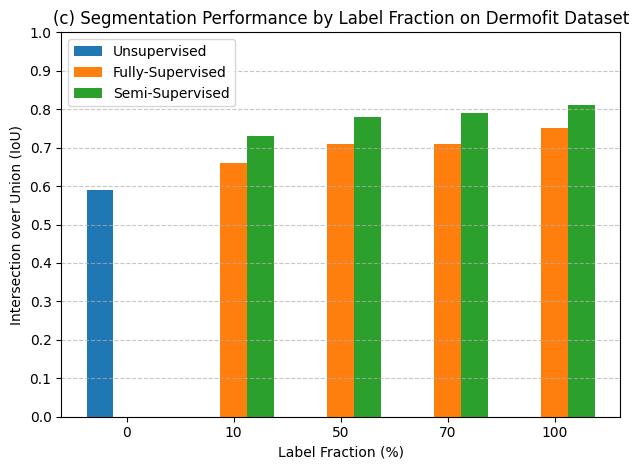}
    
    \caption{
    In this figure, we present the results for the Dermatomyositis dataset, ISIC-2017, and the Dermofit dataset in panels (a), (b), and (c), respectively. We compare the segmentation performance of Full, Semi, and Unsupervised architectures across these datasets, considering different percentages of label fraction (x-axis). Performance is evaluated using the Intersection Over Union (IoU), depicted on the y-axis, to compare results across all three datasets. IoU values range from 0 to 1.
    %In this figure, we present the results for the Dermatomyositis dataset, ISIC-2017, and the Dermofit dataset in panels (a), (b), and (c), respectively. We compare Full, Semi, and Unsupervised architectures for segmentation performance over the three datasets. We use IoU (Intersection Over Union) to compare the performance on all three datasets. The value of IoU ranges from 0 to 1. %Here, Rn34 represents ResNet-34, while ST represents Swin-U-Net. 
    The blue bar represents the performance of the unsupervised approach, PiCIE \cite{picie2021}, which, by definition, does not require any labels for fine-tuning. Consequently, we present results for PiCIE using 0\% label fractions. Remarkably, we observe that the semi-supervised approach surpasses the fully-supervised approach by requiring 50\% fewer labels per image across all three datasets.
    %Since \cite{picie2021} is an unsupervised approach, it does not require any labels for fine-tuning. Hence, we present results for PiCIE %using ResNet-34 backbone under 
    %using 0\% label fractions.  we observe that the semi-supervised approach outperforms the fully-supervised approach by using 50\% fewer per image labels on all three datasets.
    }
    \label{table: fullyvsSemi}
\end{figure}

\autoref{table: fullyvsSemi}  shows the evaluation results of these methods for the three datasets. The evaluation metric applied is intersection over union (IOU). The unsupervised approach \cite{picie2021} sets the lower bound on the segmentation performance and illustrates the difficulty of unsupervised segmentation and its ability to outperform even the state-of-the-art unsupervised approaches using a small number of annotations (10\%). Semi-supervision continuously outperformed the transfer learning across all label fractions across all datasets except the 10\% case for ISIC, where the performance is close. For all the datasets, semi-supervision with 50\% labels outperforms transfer learning from Imagenet with 100\% labels. These results show a significant performance gain by moving to machine supervision approaches instead of transfer learning from Imagenet. \autoref{fig:cls_pipeline} and \autoref{fig:seg_pipeline} show the best-performing approaches for learning from limited annotation datasets for classifications and segmentation, respectively.

\section{Discussion}\label{sec8}

% \subsection{Classification}

%We observed that the performance of the trained classifiers increased with the introduction of pretraining self-supervised objectives. Comparing the two training regimes, CASS more consistently improved performance over DINO in almost all cases under study. Studying the pixel attribution of the trained methods to understand their decision-making, we observed that CASS usually attributes the relevant pixels more than DINO. Hence, using self-supervision, i.e., additional training even with reduced human supervision, improved performance.\\

We observed a significant improvement in the performance of trained classifiers with the incorporation of pretraining self-supervised objectives, surpassing methods reliant on fully annotated data. When comparing the two training regimes, CASS consistently outperformed DINO in almost all cases under study. Further analysis into the pixel attribution of the trained methods, to understand their decision-making processes, revealed that CASS typically focuses on relevant pixels more accurately than DINO. Therefore, employing self-supervision—additional training with reduced human supervision—enhanced performance.

% \subsection{Segmentation}

%We studied three training regimes for segmentation: supervised, semi-supervised, and unsupervised. While applying the unsupervised approach requires some adaptation and hyperparameter tuning, the dermofit data set consistently shows that semi-supervised outperforms fully-supervised, regardless of labeling percentage. However, we consistently improve performance using a semi-supervised approach – wherein we used labeled and unlabeled data per batch to help the model learn better representation. The result on the dermatomyositis dataset using no labels is close as compared to the supervised and semi-supervised with 10\% labels.  Between 50\% and 70\% labeling, the performance of semi-supervised and fully-supervised merges. However, the result on the ISIC-2017 data set with 10\% labels indicates fully-supervised performs better than the semi-supervised; however as the percentage of labels increases above 10\% semi-supervised outperforms fully-supervised.   

We explored three training regimes for segmentation: supervised, semi-supervised, and unsupervised. Although implementing the unsupervised approach necessitates some adaptation and hyperparameter tuning, the Dermofit dataset consistently demonstrates that the semi-supervised approach outperforms the fully-supervised approach, irrespective of the labeling percentage. Notably, we achieved consistent performance improvements using a semi-supervised approach—wherein both labeled and unlabeled data per batch were utilized to enable the model to learn better representations. On the Dermatomyositis dataset, the performance using no labels is closely comparable to that of the supervised and semi-supervised approaches with 10\% labels. Performance between semi-supervised and fully-supervised approaches converges when labeling ranges between 50\% and 70\%. However, on the ISIC-2017 dataset with 10\% labels, the fully-supervised approach initially shows superior performance compared to the semi-supervised approach; yet, as the percentage of labels increases beyond 10\%, semi-supervised begins to outperform fully-supervised.

Additionally, unlike the unsupervised approach, the supervised and the semi-supervised approaches don't treat this problem as pixel clustering and are less prone to overfitting to the dominant pixel distribution. Since classification requires image-level identification of classes instead of pixel-level identification, segmentation is a more difficult objective. 
Yet, introducing reduced human supervision with pseudo-labels in the semi-supervised approach improved performance beyond both the supervised and unsupervised paradigms, to the point where the semi-supervised approach for segmentation outperformed fully-supervised methods while requiring 50\% fewer labels across all evaluated datasets.

% Our pipeline has been shown to surpass traditional transfer learning models, achieving superior performance with significantly fewer labeled examples.

\section{Conclusion}\label{sec13}

The primary focus of our experimental investigations revolved around two fundamental medical imaging tasks: classification and segmentation. Significantly, our findings from employing the S4MI (Self-Supervision and Semi-Supervision for Medical Imaging) framework indicate a promising shift away from traditional, purely transfer-learning-based supervised methodologies. Specifically, in the realm of classification, our empirical evidence consistently shows that self-supervised training within the S4MI framework outperforms conventional supervised approaches across both Convolutional Neural Network (CNN) and Transformer models.

Moreover, in the comparative analysis of the two self-supervised techniques, CASS demonstrates superiority by exhibiting enhanced performance in nearly all scenarios compared to DINO. Utilizing reduced supervision in segmentation has yielded favorable outcomes. Specifically, applying the semi-supervised approach, as outlined in \cite{luo2021ctbct}, has resulted in notable enhancements in performance across all label fractions for the three datasets under consideration.

A notable inference emerged from our investigation, wherein the complete elimination of supervisory signals through the utilization of unsupervised algorithms resulted in comparable performance solely in the context of the dermatomyositis dataset when compared to architectures with 10\% supervision, an observation that underscores the inherent capacity of unsupervised methodologies in distinct contextual situations.

%By making the S4MI pipeline available as open-source code

Based on a comprehensive analysis of our experimental findings, adopting machine-level supervision through the S4MI framework reduces dependence on human supervision and yields substantial advantages in both time efficiency and accuracy of medical image analysis. The findings of our study make significant empirical contributions to the fields of medical imaging and limited-supervision techniques, thereby stimulating future research in this area. The distribution of our S4MI pipeline as open-sourced code will aid other researchers by saving labeling time and improving image analysis quality. This, in turn, will facilitate the advancement of healthcare solutions, leading to improved patient care outcomes. We are optimistic that our study will act as a catalyst for meaningful dialogue and collaborative endeavors within the medical imaging community, propelling progress in this critical domain.

%Based on a comprehensive analysis of our experimental findings, adopting machine-level supervision, such as the S4MI framework we examined reduces dependence on human supervision and also yields substantial advantages in time and accuracy of medical image analysis. The findings of our study offer significant empirical contributions to medical imaging and limited-supervision techniques, thereby stimulating future investigations in this area. The distribution as open-sourced code of our  S4MI processes will support other researchers’ efforts by saving time in labeling and improving image analysis. This, in turn, will facilitate the advancement of healthcare solutions, leading to the eventual enhanced outcomes in patient care. We are optimistic that our study will serve as a catalyst for substantive discourse and cooperative endeavors within the medical imaging community, thereby propelling progress in this pivotal domain.

\section{Data Availability}
\label{data-avail}
%We have used three datasets for our experimentation - Dermatomyositis, the Dermofit and the the ISIC-2017 dataset. The Dermatomyositis dataset is a private dataset available from NYU Langone but restrictions apply to the availability of this dataset, which were used under license for the current study, and so are not publicly available. 
%Data are however available upon reasonable request and with permission of NYU Langone to Jacopo Cirrone at cirrone@courant.nyu.edu.
%To aid open-source reporduction we have carried our experiments on The Dermofit and the the ISIC-2017 dataset, both of which are openly available 
We utilized three datasets for our experiments: Dermatomyositis, Dermofit, and ISIC-2017. The Dermatomyositis dataset, while private and available from NYU Langone, is subject to restrictions and was used under license for this study; therefore, it is not publicly available. Data are however available upon reasonable request and with permission of NYU Langone to Jacopo Cirrone at cirrone@courant.nyu.edu. To facilitate open-source reproduction, we conducted our experiments on the Dermofit and ISIC-2017 datasets, both of which are publicly available at \href{https://challenge.isic-archive.com/landing/2017/}{ISIC Data repository} and \href{https://homepages.inf.ed.ac.uk/rbf/DERMOFIT/datasets.htm}{DERMOFIT Project Datasets}.

%%===========================================================================================%%
%% If you are submitting to one of the Nature Portfolio journals, using the eJP submission   %%
%% system, please include the references within the manuscript file itself. You may do this  %%
%% by copying the reference list from your .bbl file, paste it into the main manuscript .tex %%
%% file, and delete the associated \verb+\bibliography+ commands.                            %%
%%===========================================================================================%%

\bibliography{sn-bibliography}% common bib file
%% if required, the content of .bbl file can be included here once bbl is generated
%%\input sn-article.bbl

\end{document}